\documentclass[sigconf]{acmart}
\usepackage{graphicx}
\usepackage{array}
\usepackage{stackengine}
\usepackage{caption}
\usepackage{multirow}
\usepackage[
 font=large
 ]{subfig}
\usepackage[compact]{titlesec}
\usepackage[ruled,vlined]{algorithm2e}

\newcommand{\Lagr}{\mathcal{L}}
\copyrightyear{2021}
\acmYear{2021}
\setcopyright{acmcopyright}\acmConference[CODS COMAD 2021]{8th ACM IKDD CODS and 26th COMAD}{January 2--4, 2021}{Bangalore, India}
\acmBooktitle{8th ACM IKDD CODS and 26th COMAD (CODS COMAD 2021), January 2--4, 2021, Bangalore, India}
\acmPrice{15.00}
\acmDOI{10.1145/3430984.3431037}
\acmISBN{978-1-4503-8817-7/21/01}
\copyrightyear{2021}
\acmYear{2021}
\acmISBN{978-1-4503-8817-7/20/01}

\begin{document}

\title{It's LeVAsa not LevioSA!\\Latent Encodings for Valence-Arousal Structure Alignment}


\author{Surabhi S. Nath}
\authornote{Both authors contributed equally to this research.}
\affiliation{
  \institution{IIIT Delhi, India}}
  \email{surabhi16271@iiitd.ac.in}

\author{Vishaal Udandarao}
\authornotemark[1]
\affiliation{
  \institution{IIIT Delhi, India}}
\email{vishaal16119@iiitd.ac.in}

\author{Jainendra Shukla}
\affiliation{
  \institution{IIIT Delhi, India}}
  \email{jainendra@iiitd.ac.in}

\renewcommand{\shortauthors}{Nath and Udandarao, et al.}

\begin{abstract}
In recent years, great strides have been made in the field of affective computing. Several models have been developed to represent and quantify emotions. Two popular ones include (i) categorical models which represent emotions as discrete labels, and (ii) dimensional models which represent emotions in a Valence-Arousal (VA) circumplex domain. However, there is no standard for annotation mapping between the two labelling methods. We build a novel algorithm for mapping categorical and dimensional model labels using annotation transfer across affective facial image datasets. Further, we utilize the transferred annotations to learn rich and interpretable data representations using a variational autoencoder (VAE). We present ``LeVAsa'', a VAE model that learns implicit structure by aligning the latent space with the VA space. We evaluate the efficacy of LeVAsa by comparing performance with the Vanilla VAE using quantitative and qualitative analysis on two benchmark affective image datasets. Our results reveal that LeVAsa achieves high latent-circumplex alignment which leads to improved downstream categorical emotion prediction. The work also demonstrates the trade-off between degree of alignment and quality of reconstructions.
\end{abstract}



\keywords{Affective computing, representation learning, variational autoencoder, valence, arousal, circumplex model}


\maketitle

\section{Introduction}
Emotions are intrinsic characteristics of mammals, particularly overt in human behaviour \cite{darwin1998expression, panksepp2004affective, izard2013human}. Intelligent systems must employ means to incorporate emotions for a more natural interaction \cite{picard2000affective}. This surge for ``emotional intelligence'' has evolved into the field of affective computing, which by definition encompasses the creation of and interaction with machines that can sense, recognize, respond to, and influence emotions \cite{picard2002computers}. Several models of emotion have been developed over the years, which are considered as the backbone of affective computing \cite{gratch2009assessing, marsella2010computational, tracy2011four, hamann2012mapping}. Among these models, a popular choice is the Categorical Model which describes six basic discrete emotions, namely, happiness, anger, disgust, sadness, fear, and surprise \cite{ekman1971constants}. However, this model failed to capture relations between the discrete emotions. Moreover, there is a lack of consistency in the choice of these fundamental emotions \cite{ekman2011meant}. As a result, Russell \& Mehrabian \cite{russell1977evidence} developed the Dimensional Model which suggests that each emotional state can be defined in terms of Valence (pleasure of an emotion), Arousal (energy of an emotion) and Dominance (controlling nature of an emotion). 
The Dominance dimension is commonly ignored since the valence- arousal (VA) dimensional model was shown to possess adequate reliability, convergent validity, and discriminant validity \cite{russell1989affect}. This led to the conceptualization of the Circumplex Model to represent affective states as a circle in a 2D bipolar VA space \cite{russell1980circumplex}. The VA variables are typically considered independent \cite{feldman1998independence}. 
\\
\textbf{Motivation}:
The existence of different models of emotions result in a range of possible annotation strategies for affective data \cite{fabian2016emotionet, nicolaou2010audio, nicolaou2011continuous, lucey2010extended, kossaifi2017afew}. This poses two challenges: (i) building deep models on affective data, and (ii) drawing collective insights from multiple datasets having potentially different formats of annotations \cite{de2019joint}. 

In this paper, we present a novel algorithm for mapping annotations of the Categorical Model to those of the Dimensional Model through annotation transfer across affective facial image datasets. The subsequent step following annotation mapping is to obtain meaningful representations. With the increased use of deep neural networks and generative models, there have been significant advances in emotion modelling and affective computing \cite{han2019adversarial, rouast2019deep, jolly2019universal}. Variational Autoencoders (VAEs) \cite{kingma2013auto} are known to yield disentangled latent representations and generate new data samples \cite{dis1, dis2, dis3}. They have been used extensively in affective computing to represent text, audio, image and electroencephalography (EEG) data \cite{wu2019semi, latif2017variational}. Applying VAEs on affective facial images to obtain disentangled image representations can (i) provide high quality feature representations for downstream tasks \cite{bengio2013representation, peters2017elements}, and (ii) serve applications like facial editing and data augmentation \cite{lindt2019facial}. 
In our study, we obtain interpretable features by aligning the latent space of a VAE with the VA space. We show that this improves performance in downstream tasks of affect classification and regression, as demonstrated on two benchmark affective image datasets.\vspace{2mm}
\\
Our major contributions are (i) an annotation transfer algorithm for label transfer between Categorical and Dimensional models of emotion and (ii) a regularised VAE model ``LeVAsa'' (Latent Encodings for Valence-Arousal Structure Alignment) that yields an interpretable latent space with an implicit structure aligned with the VA space.


\section{Methods}
\label{method}
Here we present our annotation transfer algorithm and VAE model architectures. Our code and models are publically available \footnote{\textcolor{magenta}{\href{https://github.com/vishaal27/LeVAsa}{https://github.com/vishaal27/LeVAsa}}}.
\subsection{Annotation Transfer Algorithm}
For the task of annotation transfer between Categorical and Dimensional emotion models, we use an external reference dataset ($D_r$) containing both discrete categorical emotion labels ($l_i \in \{e_1, e_2, ...,e_n\}$, where $e_1, e_2, ...,e_n$ are the n discrete emotional labels)  and valence, arousal values ($v_i \in [llim_v, ulim_v], a_i \in [llim_a, ulim_a]$, where $llim_v, ulim_v, llim_a, ulim_a$ are the lower and upper limits for valence and arousal values respectively). Each data sample $x_i \in D_r$ thus has an emotion label $l_i$, a valence value $v_i$ and an arousal value $a_i$. $D_r$ serves as the standard based on which continuous or discrete VA values can be sampled for data points in a working dataset ($D$) with only emotion labels (Algorithm \ref{algo}, Line \ref{line4}), or conversely, the most likely emotion labels can be obtained for data points in dataset $D^{\prime}$ with only VA tuples (Algorithm \ref{algo}, Line \ref{line6}). The ellipse sizes (on average 3\% of total area) ensure that the sampled VA values allow both sufficient variability and consistency within emotion classes.

\begin{algorithm}[!h]
\LinesNumbered
\SetKwData{Left}{left}\SetKwData{This}{this}\SetKwData{Up}{up}
\SetKwFunction{Union}{Union}\SetKwFunction{FindCompress}{FindCompress}
\SetKwInOut{Input}{Input}\SetKwInOut{Output}{Output}
\Input{reference dataset $D_r$, discrete categorical emotion labels $[l_1, l_2, ... l_{|D_r|}]$, VA values $[v_1, v_2, ... v_{|D_r|}]$ and $[a_1, a_2, ... a_{|D_r|}]$, working dataset $D$ with discrete emotion labels, working dataset $D'$ with VA tuples}
\Output{VA values for the working dataset $D$, discrete emotion labels for the working dataset $D'$}
\BlankLine
{Partition each sample $x_i \in D_r$ into $n$ groups $(g_1, g_2,..., g_n)$ based on discrete emotion labels $l_i \in (e_1, e_2,.. e_n)$}\\
{For each group $g_j$, $j \in \{1, 2,..., n\}$, obtain the mean valence $\mu_{vj}$, standard deviation valence $\sigma_{vj}$, mean arousal $\mu_{aj}$, standard deviation arousal $\sigma_{aj}$ values}\\
{Generate ellipses $c_j$, $j \in \{1, 2.., n\}$ for group $g_j$ representing emotion $e_j$ with centre $(\mu_{vj}, \mu_{aj})$, semi major axis $\sigma_{vj}$ and semi minor axis $\sigma_{aj}$}\\
{To obtain VA values for data point $x_k$ in $D$ with label $l_k$, sample $(v_k, a_k)$ from ellipse $c_k$ as: $v_k = x\sigma_{vk} + \mu_{vk}$, $a_k = y\sigma_{ak} + \mu_{ak}$, where $x = \sqrt{r} \cos\theta$, $y = \sqrt{r}\sin\theta$ and $r \sim [0,1]$, $\theta \sim [0, 2\pi]$ \label{line4}}\\
{Convert $(v_k, a_k)$ to discrete values by scaling and rounding-off if desired}\\
{To obtain emotion label for sample $x_k$ in $D’$ with VA $(v_k, a_k)$, find ellipse $c_k$ with centroid at least Euclidean distance from $(v_k, a_k)$, and assign $e_k$ as most likely emotion \label{line6}}
 \caption{Annotation transfer algorithm}
 \label{algo}
\end{algorithm}

\renewcommand{\thefigure}{1}
\subsection{VAE model architectures}
\begin{figure}
    \centering
    \includegraphics[scale=0.35, trim=0 30 0 80]{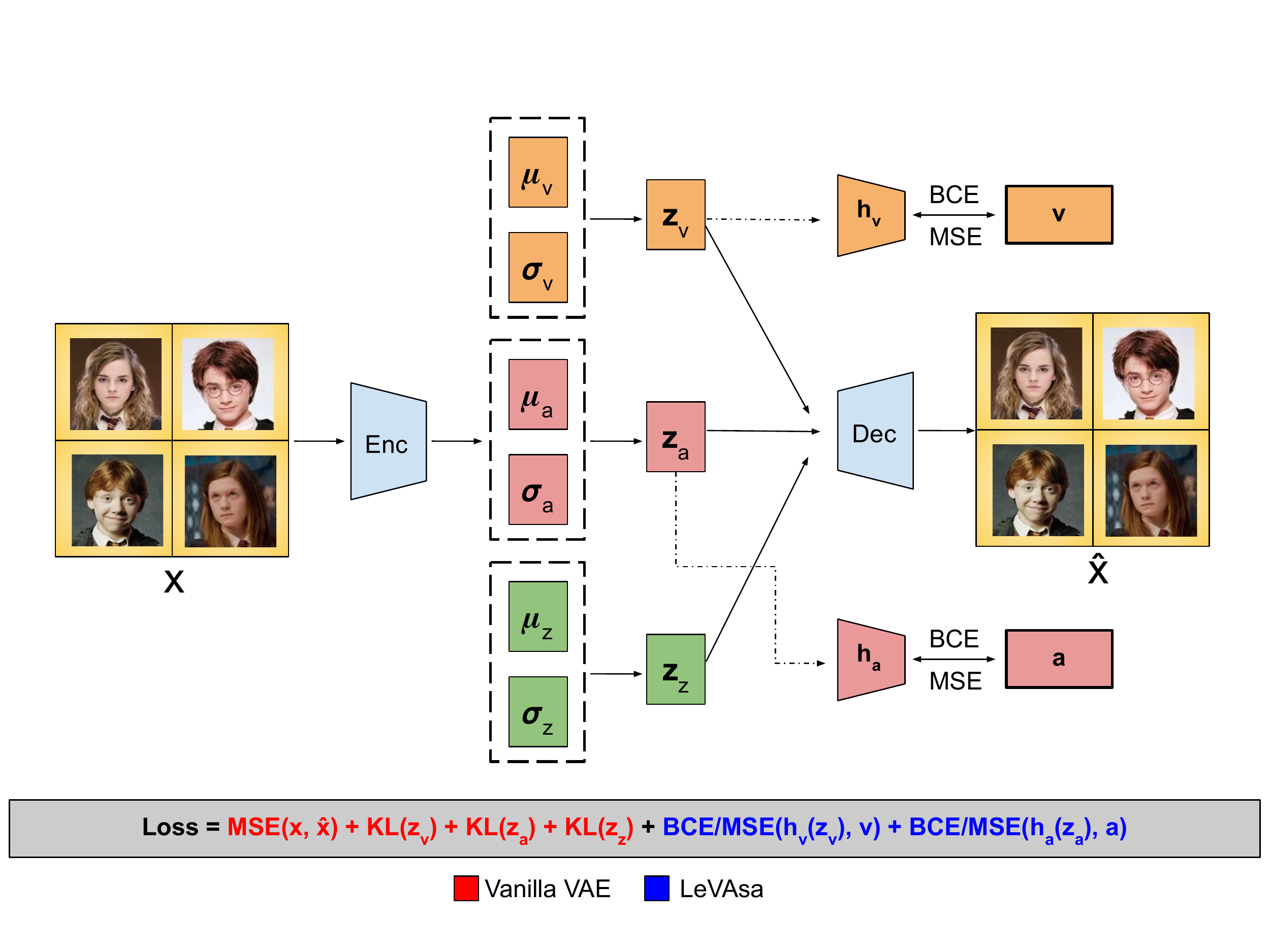}
    \caption{Model Architecture}
    \label{arch}
\end{figure}

We train a generative model with an interpretable latent space with an implicit structure given a raw distribution of affective face images. We employ variational autoencoder based models because of their simple training protocols and structured inductive priors. We compare two VAE models, Vanilla VAE and LeVAsa. This comparison is justified because: (1) No baseline VA structured latent model exists, and (2) the Vanilla VAE features being unbiased (owing to no explicit supervision) may perform better on the downstream tasks. The latent space for both models was constructed to comprise of three chunks.
Figure \ref{arch} depicts our model architectures.

For the Vanilla VAE, no explicit alignment was imposed on the latent space, whereas for LeVAsa, we take inspiration from recent work \cite{dis4, dis6} and model the latent space as follows:
\begin{itemize}
	\item $Z_v$ -- subspace consisting of valence attributes $z_v$ that learn to encode the valence features of image samples
	\item $Z_a$ -- subspace consisting of arousal attributes $z_a$ that learn to encode the arousal features of image samples
	\item $Z_z$ -- subspace consisting of other miscellaneous generative attributes $z_z$ that are required for high-fidelity reconstruction of the input data distribution.
\end{itemize}

Given a dataset of $N$ affective images $X = \{x_1, x_2,..., x_N\}$, our VAE backbone consists of encoder $f_\theta$ and decoder $g_\phi$ given by:
\begin{gather*}
z_{v_i}, z_{a_i}, z_{z_i} = f_\theta(x_i), x_i \in X\\
\hat x_i = g_\phi(z_{v_i}, z_{a_i}, z_{z_i})\\
z_{v_i} \in {Z_v}, z_{a_i} \in Z_a, z_{z_i} \in Z_z
\end{gather*}

We train the Vanilla VAE with a simple reconstruction loss along with a modified Kullback-Leibler (KL) loss (Eq. \ref{eqn}). We induce a $\mathbf{\mathcal{N}(\boldsymbol{0}, \boldsymbol{I})}$ prior on all three attributes $z_v$, $z_a$ and $z_z$.
\begin{gather}
 \Lagr_{KL} = \sum_{z \in \{z_v, z_a, z_z\}}KL\Big(f_{\theta}(z|x) \| \mathcal{N}(\boldsymbol{0}, \boldsymbol{I}))\Big) \footnotemark
 \label{eqn}
\end{gather}
We employ the same backbone Vanilla VAE architecture for the LeVAsa model with two major modifications:
\begin{enumerate}
	\item \textit{Projection Heads}: We use two non-linear projection heads ${h}_v$ and ${h}_a$ which map the encoded valence and arousal representations $z_v$ and $z_a$ to the valence and arousal label space (giving label representations $r_v$ and $r_a$). The projections obtained are represented as follows:
	\begin{equation*}
	    r_{v_i} = h_v(z_{v_i}), r_{a_i} = h_a(z_{a_i})
	\end{equation*}
\item \textit{VA-regularization loss}: To impose an explicit alignment of the $z_v$ and $z_a$ attributes with the VA ground truth factors, we introduce a VA-regularization loss as follows:
\begin{gather}
\label{Lc}
\Lagr_{C} = \sum_{i=0}^{N}\Big({\Lagr (r_{v_i}, v_i) + \Lagr (r_{a_i}, a_i)}\Big)
\end{gather}
where $\Lagr$ takes the form of MSE for continuous and BCE for discrete annotation types. MSE/BCE are design choices and can be replaced by suitable likelihood-based loss functions.
\end{enumerate}
\noindent
The overall optimization objective for the LeVAsa model is:
\begin{gather}
    \Lagr_{total} = \Lagr_{R} + \lambda_{KL} \Lagr_{KL} + \lambda_{C} \Lagr_{C}
\end{gather}
where $\lambda_{KL}$ and $\lambda_C$ are hyperparameters.

\subsection{Datasets}
We use the following datasets in our experiments.
\subsubsection*{Annotation Transfer: AffectNet}
\begin{itemize}
    \setlength{\itemsep}{1pt}
    \setlength{\parskip}{0pt}
    \setlength{\parsep}{0pt}
    \item \textbf{AffectNet} \cite{mollahosseini2017affectnet} is the largest facial expression dataset, with over 420,000 annotated images and contains both continuous VA annotations and discrete emotional labels. The dataset also incorporates a wide diversity in gender, age and ethnicity, hence is an ideal choice for the reference dataset in the annotation transfer algorithm (Algorithm \ref{algo}). The generated ellipses are shown in Figure \ref{ellipses}.
\end{itemize}

\subsubsection*{Model Training: IMFDB, AFEW-VA}
\begin{itemize}
    \setlength{\itemsep}{1pt}
    \setlength{\parskip}{0pt}
    \setlength{\parsep}{0pt}
  \item \textbf{IMFDB} \cite{setty2013indian} contains around 34,000 annotated zoomed-in facial images of 100 Indian actors, with only emotional labels and no VA supervision. Continuous and discrete VA supervision for IMFDB is obtained from annotation transfer using AffectNet. This is particularly well suited due to the similar nature of images in IMFDB and AffectNet datasets.
  \item \textbf{AFEW-VA} \cite{kossaifi2017afew} on the other hand, contains around 24,000 annotated images from videos of real world scenes of approximately 600 actors with only discrete VA values and no discrete emotional labels. 
\end{itemize}
The different nature of IMFDB and AFEW-VA datasets allow us to analyse and compare model performance based on different factors including image type (zoomed in faces/video scenes) and annotation type (discrete VA supervision/continuous VA supervision).
\vspace{-4mm}
\renewcommand{\thefigure}{2}
\begin{figure}[H]
    \centering
  \stackunder{\includegraphics[scale=0.25]{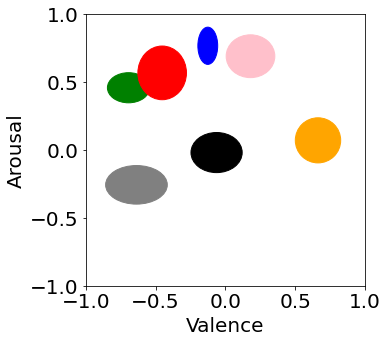}}{\textbf{\fontsize{9}{20}\selectfont{\textcolor{white}{s}(a) Continuous}}}\hspace{1mm}%
  \stackunder{\includegraphics[scale=0.25]{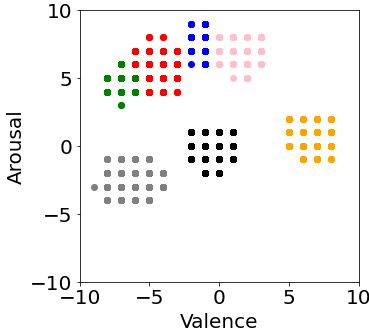}}{\textbf{\fontsize{9}{20}\selectfont{\textcolor{white}{sp}(b) Discrete}}}
  \subfloat{\raisebox{5ex}
      {\includegraphics[scale=0.25]{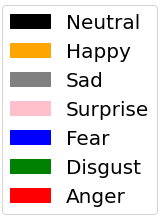}}}
  \caption{Ellipses from AffectNet for annotation transfer}
  \label{ellipses}
\end{figure}

\vspace{-5mm}
\section{Experiments}
\label{exp}

We evaluate model performance through a series of qualitative and quantitative experiments. This enables comparisons based on three aspects: (i) architecture (Vanilla VAE vs LeVAsa), (ii) dataset (IMFDB vs AFEW-VA), and (iii) nature of annotations (Continuous VA vs Discrete VA). Altogether, we train five models: (i) Vanilla VAE on IMFDB, (ii) LeVAsa on IMFDB with continuous VA annotations, (iii) LeVAsa on IMFDB with discrete VA annotations, (iv) Vanilla VAE on AFEW-VA, (v) LeVAsa on AFEW-VA with discrete VA annotations.

\renewcommand{\thefigure}{4}
\begin{figure*}[!t]
\begin{minipage}{.2\linewidth}
\centering
\subfloat[IMFDB Vanilla VAE]{
\includegraphics[scale=.3]{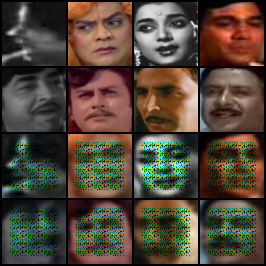}}
\end{minipage}%
\begin{minipage}{.2\linewidth}
\centering
\subfloat[IMFDB LeVAsa Cont.]{
\includegraphics[scale=.3]{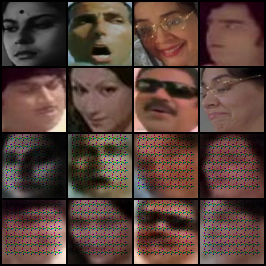}}
\end{minipage}%
\begin{minipage}{.2\linewidth}
\centering
\subfloat[IMFDB LeVAsa Discrete]{
\includegraphics[scale=.3]{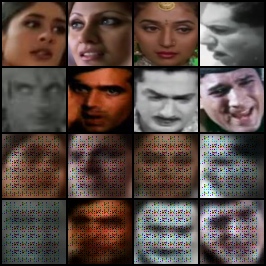}}
\end{minipage}%
\begin{minipage}{.2\linewidth}
\centering
\subfloat[AFEW-VA Vanilla VAE]{
\includegraphics[scale=.3]{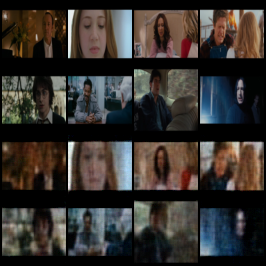}}
\end{minipage}%
\begin{minipage}{.2\linewidth}
\centering
\subfloat[AFEW-VA LeVAsa \textcolor{white}{dumbesss}]{
\includegraphics[scale=.3]{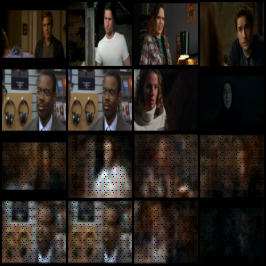}}
\end{minipage}%
\caption{VAE Reconstruction from the five models}
\label{recon}
\vspace{-10pt}
\end{figure*}

\subsection{Latent-Circumplex Alignment}
We measure the alignment of LeVAsa's $Z_v \cup Z_a$ latent space with the VA ground truths using normalized Euclidean and Manhattan distance metrics for continuous annotations, and Cross Entropy measure for discrete annotations. This helps quantify the degree of latent-circumplex alignment. For the Vanilla VAE, we determine the $z_v$ and $z_a$ chunks heuristically by considering the two latent chucks which aligned best with the corresponding valence and arousal ground truths. Further, we reduce the dimensionality of the $z_v$ and $z_a$ latent chunks and plot them alongside the ground truth to replicate the circumplex representation.

It is found that LeVAsa outperformed Vanilla VAE for both continuous and discrete annotations (Table \ref{align}). This clearly exhibits the superior latent-circumplex alignment achieved by LeVAsa. For discrete annotations, in case of AFEW-VA, the difference between the cross entropy measures of the Vanilla VAE and LeVAsa is greater than in case of IMFDB. This could be attributed to the different image types in both datasets. The circumplex plots (Figure \ref{circum}) for LeVAsa reveal reduced variance and increased alignment with true labels. This validates the quantitative results in Table \ref{align}.

\renewcommand{\thetable}{1}
\setlength{\tabcolsep}{6pt} 
\renewcommand{\arraystretch}{1}
\begin{table}[H]
  \caption{Alignment}  \label{tab:discrete_alignment}
  \label{align}
  \begin{tabular}{ccccccc}  \toprule
    \multirow{2}{*}{IMFDB} & \multicolumn{2}{c}{Valence} & \multicolumn{2}{c}{Arousal} & \multicolumn{2}{c}{Combined}\\
    & MSE & MAE & MSE & MAE & MSE & MAE\\
    \midrule
     Vanilla VAE & 1.83 & 0.29 & 1.49 & 0.26 & 3.31 & 0.55\\
     LeVAsa & 0.14 & 0.14 & 0.06 & 0.09 & 0.2 & 0.23\\
  \bottomrule\\[-6pt]
  \multicolumn{7}{c}{\textbf{(a) Continuous}}
\end{tabular}
\end{table}
\vspace{-5mm}
\begin{table}[H]
  \begin{tabular}{ccc}  \toprule
    Model & IMFDB & AFEW-VA\\
    \midrule
    Vanilla VAE & 8.9 & 8.9\\
    LeVAsa & 6.63 & 2.54\\
  \bottomrule\\[-6pt]
  \multicolumn{3}{c}{\textbf{(b) Discrete}}
\end{tabular}
\end{table}
\vspace{-6mm}
\renewcommand{\thefigure}{3}
\begin{figure}[H]
\centering
    \subfloat{\includegraphics[scale=0.3]{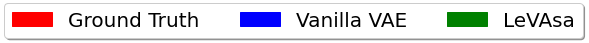}}
    \\
    \addtocounter{subfigure}{-1}
  \subfloat[IMFDB Cont.]{\includegraphics[width=1.107in, , trim=0 0 0 20]{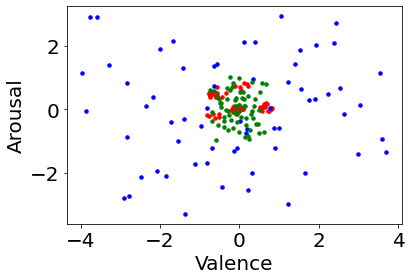}}
    \subfloat[IMFDB Disc.]{\includegraphics[width=1.14in, trim=0 0 0 20]{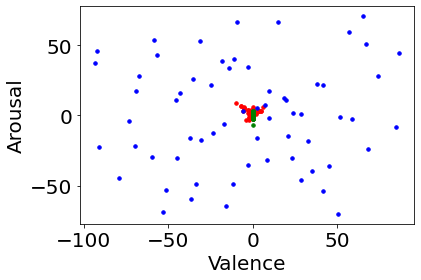}}
  \subfloat[AFEW-VA]{\includegraphics[width=1.135in, , trim=0 0 0 20]{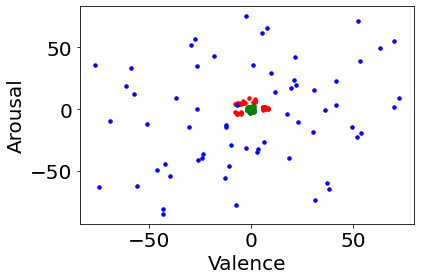}}
  \caption{Circumplex Representation}
  \label{circum}
  \vspace{-2mm}
\end{figure}

To gain further insights, we assess the regressive power of $z_v$ and $z_a$ by their ability to predict the corresponding VA ground truths. We use Multi Layer Perceptron (MLP) Regression. This analysis applies to continuous annotations hence was conducted only on the models trained on IMFDB dataset with continuous VA values. 
\vspace{-1pt}
\renewcommand{\thetable}{2}
\begin{table}[H]
  \caption{VA Regressive Power}
  \label{reg}
  \begin{tabular}{cccccc}  \toprule
    Axis & Model & MSE & MAE & EV & $R^2$\\
    \midrule
    Valence & \begin{tabular}[c]{@{}c@{}}Vanilla VAE\\ LeVAsa\end{tabular} & \begin{tabular}[c]{@{}c@{}}0.256\\ 0.251\end{tabular} & \begin{tabular}[c]{@{}c@{}}0.420\\ 0.414\end{tabular} & \begin{tabular}[c]{@{}c@{}}-0.011\\ 0.016\end{tabular} & \begin{tabular}[c]{@{}c@{}}-0.012\\ 0.015\end{tabular}\\
    \midrule
    Arousal & \begin{tabular}[c]{@{}c@{}}Vanilla VAE\\ LeVAsa\end{tabular} & \begin{tabular}[c]{@{}c@{}}0.092\\ 0.074\end{tabular} & \begin{tabular}[c]{@{}c@{}}0.242\\ 0.224\end{tabular} & \begin{tabular}[c]{@{}c@{}}-0.022\\ 0.095\end{tabular} & \begin{tabular}[c]{@{}c@{}}-0.048\\ 0.086\end{tabular}\\    
  \bottomrule
\end{tabular}
\end{table}

It is observed that the MSE and MAE values computed for LeVAsa were lower by 2.25\% and 1.42\% as compared to Vanilla VAE for valence, and lower by 19.13\% and 7.18\% as compared to Vanilla VAE for arousal (Table \ref{reg}). Furthermore, the goodness of fit metrics (explained variance and $R^2$) showed better performance in the case of LeVAsa. These results further strengthen our hypothesis.

\subsection{Categorical Emotion Predictive Power}
We predict the discrete emotion labels using different combinations of latent representations obtained from Vanilla VAE and LeVAsa (Table \ref{disc_emotion_pred}). Due to lack of discrete emotion labels in the AFEW-VA dataset, it was excluded from this analysis. We randomized the data splits across Continuous and Discrete experiments to ensure an unbiased setup. Model performance is evaluated using classification accuracy. We utilize a simple one-layered MLP to ensure that the accuracy is a direct measure of representation quality and not influenced by the complexity of the classifier.

It is seen that LeVAsa has significantly better predictive power as compared to the Vanilla VAE. Moreover, for LeVAsa, the VA chunks alone are more informative in emotion prediction as compared to $z_v \oplus z_a \oplus z_z$ chunks altogether. Also, the improvement in classification accuracy by employing LeVAsa in place of Vanilla VAE can be compared under the continuous and discrete settings. This reveals that LeVAsa representations from the model trained with discrete annotations and BCE loss (Eq. \ref{Lc}) proves to be better at classifying emotion labels. This is due to the discrete nature of emotion labels which correlate well with the model representations.

\renewcommand{\thetable}{3}
\begingroup
\setlength{\tabcolsep}{3pt} 
\renewcommand{\arraystretch}{1} 
\begin{table}
  \caption{Categorical emotion predictive power using vanilla VAE and LeVAsa models. All reported scores are accuracies ($\oplus$ represents vector concatenation)}
  \label{disc_emotion_pred}
  \begin{tabular}{ccccc}
    \toprule
    Annotation & Chunk & Vanilla VAE & LeVAsa &  Difference=\\
    Type & Combination & (V) & (L) & L - V (in \%)\\
    \midrule
    & $z_v$ & 0.29 & 0.36 & 7\\
    & $z_a$ & 0.32 & 0.35 & 3\\
    Continuous & $z_z$ & 0.32 & 0.36 & 4\\
    & $z_v \oplus z_a$ & 0.32 & 0.38 & 6\\
    & $z_v \oplus z_a \oplus z_z$ & 0.29 & 0.33 & 4\\
    \midrule
    & $z_v$ & 0.30 & 0.35 & 5\\
    & $z_a$ & 0.27 & 0.30 & 3\\
    Discrete & $z_z$ & 0.24 & 0.30 & 6\\
    & $z_v \oplus z_a$ & 0.25 & 0.33 & 8\\
    & $z_v \oplus z_a \oplus z_z$ & 0.26 & 0.30 & 4\\
  \bottomrule
\end{tabular}
\end{table}

\subsection{Reconstruction Quality}

VAE models are prone to posterior collapse, producing unreliable reconstructions \cite{he2019lagging, rybkin2020simple}. Thus, along with analyses of the latent representations, we study the quality of reconstructions (Figure \ref{recon}).

It is observed that the quality of the reconstructed faces is slightly compromised in the case of LeVAsa as compared to Vanilla VAE. This can be attributed to the slightly higher variance of the learnt LeVAsa decoding distribution \cite{dis3, alemi2018fixing}. By Shannon’s rate-distortion theory \cite{berger2003rate}, there is a trade-off between the distortion (reconstruction quality) and rate (representation quality). Since we are imposing an explicit compression bottleneck on the latent space, it is expected that the reconstruction quality is slightly compromised in order to achieve better interpretability of latent representations.

\section{Conclusion}
\label{conc}
In this paper, we have developed an annotation-transfer algorithm for mapping between Categorical and Dimensional emotion model annotations. Using them, we generated interpretable image features with a VA-regularized VAE model called LeVAsa. 
We conducted a series of evaluation tasks to verify and validate our experiments and compare performance based on three factors: (i) architecture (Vanilla VAE vs LeVAsa), (ii) dataset (IMFDB vs AFEW-VA), and (iii) nature of annotations (Continuous VA vs Discrete VA). The results showed that the LeVAsa model obtains robust and interpretable representations enabling improved downstream affective task performance. In the future, we hope to extend the annotation-transfer algorithm to action-unit annotations, and test the expressiveness of the representations by performing latent traversals for data augmentation and facial editing.

\begin{acks}
This work was supported by the Infosys Center for Artificial Intelligence at IIIT Delhi, India.
\end{acks}

\bibliographystyle{ACM-Reference-Format}





\end{document}